\documentclass[runningheads]{llncs}
\usepackage{graphicx}

\usepackage{tikz}
\usepackage{comment}
\usepackage{amsmath,amssymb} 
\usepackage{color}

\begin{document}
\pagestyle{headings}
\mainmatter
\def\ECCVSubNumber{1}  

\title{Reinforcement Learning for Improving Object Detection} 

    \titlerunning{ObjectRL}
    %
    \author{Siddharth Nayak
    \and
    Balaraman Ravindran
    }
    \authorrunning{S. Nayak et al.}
    %
    \institute{Indian Institute of Technology Madras, Chennai TN, India\\
    \email{siddharthnayak98@gmail.com, ravi@cse.iitm.ac.in}}
\maketitle
\newif\iflogvar
\logvartrue
\begin{abstract}
The performance of a trained object detection neural network depends a lot on the image quality. Generally, images are pre-processed before feeding them into the neural network and domain knowledge about the image dataset is used to choose the pre-processing techniques. In this paper, we introduce an algorithm called ObjectRL to choose the amount of a particular pre-processing to be applied to improve the object detection performances of pre-trained networks. The main motivation for ObjectRL is that an image which looks good to a human eye may not necessarily be the optimal one for a pre-trained object detector to detect objects.
\end{abstract}

\section{Introduction}

With the advent of convolutional neural networks, object detection in images has improved significantly giving rise to several object detection algorithms like YOLO \cite{DBLP:journals/corr/RedmonDGF15}, SSD \cite{DBLP:journals/corr/LiuAESR15}, etc. Most object detection networks work with raw image pixels as inputs. The networks are highly nonlinear in nature and thus the output predictions depend a lot on the image parameters like brightness, contrast, etc. \cite{937690,6521924,Osadchy2004EfficientDU,6115698}. In real-world scenarios, camera parameters like the shutter-speeds, gains, etc. with which the images are taken, matter a lot in the performance of an object detection network. A photographer changes a lot of parameters like the shutter speed, voltage gains, etc. \cite{5765998} while capturing images according to the lighting conditions and the movements of the subject. In autonomous navigation, robotics, etc. there are several instances where the lighting conditions and the subject speed changes. In these cases, using fixed shutter speed and voltage-gain values would result in an image which would not be conducive for object detection. Most cameras rely on the built-in auto-exposure algorithms to set the exposure parameters of the camera. Although the images obtained from these auto-exposure algorithms may be \textit{pleasing} to a human eye, they may not be the best image to perform object detection on. Also, most of the object detection networks are trained using images from a dataset which are captured either by using a single operation mode \cite{Geiger2013IJRR} or no control over the parameters of the camera \cite{imagenet_cvpr09,Agustsson_2017_CVPR_Workshops,huiskes08}. Thus, a pre-trained network may have a larger affinity towards images captured with similar parameters as the ones in the dataset it was trained on.\\

 To tackle the problem of sudden variations in the photography conditions, we propose to train a Reinforcement Learning (RL) agent to digitally transform images in real-time such that the object detection performance is maximised. Although we perform experiments with digital transformations, this method can ideally be extended to choose the camera parameters to capture the images by using the image formation model proposed by Hassinoff et al. \cite{5540167}. We train the model with images which are digitally distorted, for example: changing brightness, contrast, color, etc. It should be noted that we do not necessarily want the agent to recover the original image. \\

The claimed contribution of the paper is a Deep RL methodology called \textit{ObjectRL} (Object Reinforcement Learning) to change the image digitally with rewards based on the performance of a pre-trained object detector on the agent-transformed image. An overview of the related work is provided in the next section. 
The proposed method for \textit{ObjectRL} is described in detail in Section \ref{section:model} and the experiments to validate the hypotheses along with results are provided in Section \ref{section:experiments} and Section \ref{section:results} respectively.

\begin{figure}[!t]
    \centering
    \includegraphics[scale=0.4]{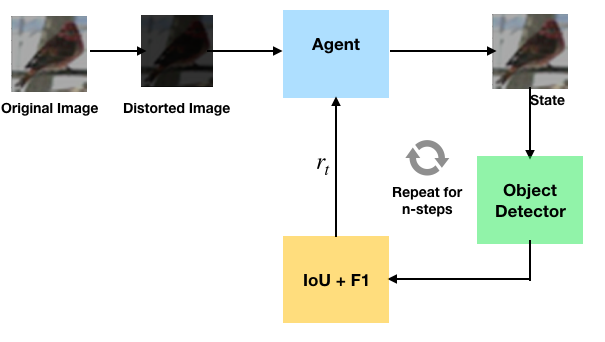}
    \caption{The overall training procedure for \textit{ObjectRL}. The image is randomly distorted to simulate the bad images. An episode can be carried out for $n$ steps which we set to 1 for training stability. Thus, the agent has to take a single action on each image.}
    \label{fig:flow}
\end{figure}

\section{Related Works} \label{section:related}
We briefly review the literature and the existing methods related to image modifications for object detection improvement. 

Bychkovsky et al. \cite{5995332} present a dataset of input and retouched image pairs called MIT-Adobe FiveK, which was created by professional experts. They use this dataset to train a supervised model for color and tone adjustment in images. The main motive of this work is not inclined towards improving object detection but is more focused towards training a model to edit an image according to the user preferences.

In \cite{DBLP:journals/corr/WuT17} the authors create a dataset of images taken with different combinations of shutter speeds and voltage gains of a camera. They create a performance table which is a matrix of mean average precision (mAP) for detection of objects in images taken with different combinations of shutter speed and gains. To choose the optimal parameters to capture images, they propose to choose the combination which gives the maximum precision. One of the problems with this method is that a dataset with images taken with different combinations of shutter speeds, voltage gains and illuminations has to be manually annotated with bounding boxes around the objects which is quite time-consuming. Also, the dataset consists of images with static objects. Thus, the effect of changing shutter speed is just on the overall brightness of the image. But one of the main reasons for changing shutter speed while capturing images is to increase (for artistic purposes) or (preferably) decrease motion blur in the moving objects.

In \cite{DBLP:journals/corr/abs-1804-04450} the authors propose a reinforcement learning based method to recover digitally distorted images. The authors model the agent to take actions sequentially by choosing the type of modification (brightness, contrast, color saturation, etc.). The main motive of this model is to recover back the distorted images. The reward for the agent is the difference of mean square difference of the images at the current time step and the previous time step. This work is quite different from our \textit{ObjectRL} model as our main motive is to maximise the object detection performance of a pre-trained detector.

Reinforcement Learning has been used in conjunction with computational photography in recent works by Yang et al. \cite{DBLP:journals/corr/abs-1803-02269} and Hu et al. \cite{10.1145/3181974} where the authors train RL agents to either capture images or post-process images in such a way that the resultant image is \textit{visually pleasing}. The agent gets a reward from the users according to their preferences of exposures on cameras in the former one whereas in the later one the agent receives a reward based on the discriminator loss of a Generative Adversarial Network \cite{10.5555/2969033.2969125}.  

Another area of research orthogonal to ours is using reinforcement learning to obtain region proposals for object-detection and object-localization\cite{DBLP:journals/corr/MatheS14,DBLP:journals/corr/abs-1810-10325,DBLP:journals/corr/CaicedoL15,7780685}. In these works, the main motivation is to make the agent focus its attention toward candidate regions to detect objects by sequentially shifting the proposed region and rewarding the agent according to the Intersection over Union $(IoU-$explained in Section \ref{section:IOU}).

\section{Background}\label{section:background}
    \subsection{Reinforcement Learning} \label{section:RL}
    Reinforcement learning (RL) tries to solve the sequential decision problems by learning from trial and error. Considering the standard RL setting where an agent interacts with an environment $\mathcal{E}$
    over discrete time steps. In the time step $t$, the agent receives a state $s_t \in \mathcal{S}$ and selects an
    action $a_t \in \mathcal{A}$ according to its policy $\pi$, where $\mathcal{S}$ and $\mathcal{A}$ denote the sets of all possible states
    and actions respectively. After the action, the agent observes a scalar reward $r_t$ and receives
    the next state $s_{t+1}$. The goal of the agent is to choose actions to maximize the cumulative sum of rewards over time. In other words,
    the action selection implicitly considers the future rewards. The discounted return is defined as $R_t = \sum_{\tau=t}^{\infty}\gamma^{\tau-t}r_{\tau}$, where $\gamma \in [0, 1]$ is a discount factor that trades-off the importance of recent
    and future rewards.

    RL algorithms can be divided into two main sub-classes: Value-based and Policy-based methods. In value-based methods, values are assigned to states by calculating an expected cumulative score of the current state. Thus, the states which get more rewards, get higher values. In policy-based methods, the goal is to learn a map from the states to actions, which can be stochastic as well as deterministic. A class of algorithms called actor-critic methods \cite{NIPS1999_1786} lie in the intersection of value-based methods and policy-based methods, where the critic learns a value function and the actor updates the policy in a direction suggested by the critic. 
    
    \textbf{Proximal Policy Optimization (PPO)}:  We use PPO \cite{DBLP:journals/corr/SchulmanWDRK17} which is a type of actor-critic method for optimising  the RL agent. One of the key points in PPO is that it ensures that a new update of the current policy does not change it too much from the previous policy. This leads to less variance in training at the cost of some bias, but ensures smoother training and also makes sure the agent does not go down an unrecoverable path of taking unreasonable actions. PPO uses a clipped surrogate objective function which is a first order trust region approximation. The purpose of the clipped surrogate objective is to stabilize training via constraining the policy changes at each step. 
    
    \subsection{Object Detection}\label{section:IOU}
    Object recognition is an essential research direction in computer vision. Most of the successful object recognition algorithms use deep convolutional neural networks which are trained to give the co-ordinates of the bounding boxes around the objects. To decide whether an object is detected or not, we use the Intersection over Union (IoU) criteria. Intersection over Union is the ratio of area of overlap and area of union of the predicted and the ground truth bounding boxes. Let $p$ be the predicted box, and $g$ be the ground truth box for the target object. Then, $IoU$ between $p$ and $g$ is defined as $IoU(p,g)=Area(p\cap g)/Area(p\cup g)$. Generally, if $IoU> 0.5$ an object is said to be a True-Positive. 
    
    \subsection{Image Distortions}\label{section:img_distortions}
        Different  parameters  of  an  image  like  brightness,  contrast and color can be changed digitally. We describe the formulae used to transform the pixel intensity $(I)$ values at  the  co-ordinates(x,y). We assume distortion factor $\alpha\geq0$
        \begin{itemize}
            \item Brightness: The brightness of an image   can be changed by a factor $\alpha$ as follows:\\ $I(x,y)\gets \min (\alpha I(x,y),255))$
            \item Color: The color of an image is changed by a factor $\alpha$ as follows: We evaluate the gray-scale image as:\\
            $gray= (I(r) +I(g) +I(b))/3$, where I(r), I(g) and I(b) are the R, G \& B pixel values respectively.\\
            $ I(x,y)\gets \min(\alpha I(x,y) + (1 - \alpha)gray(x,y),255)$
            \item Contrast: The contrast in an image is changed by a factor $\alpha$ as follows:\\
            $\mu_{gray}=mean(gray)$\\
            $I(x,y)\gets\min(\alpha I(x,y) + (1 - \alpha)\mu_{gray},255)$
        \end{itemize}
        
\section{Model}\label{section:model}
    Given an image, the goal of \textit{ObjectRL} is to provide a digital transformations which would be applied to the input image. This transformed image should extract maximum performance (F1 score) on object detection when given as an input to a pre-trained object detection network.

    \subsection{Formulation}
    We cast the problem of image parameter modifications as a Markov Decision Process (MDP) \cite{Puterman:1994:MDP:528623} since this setting provides a formal framework to model an agent that makes a sequence of decisions. Our formulation considers a single image as the state. To simulate the effect of \textit{bad} images as well as increase the variance in the images in a dataset, we digitally distort the images. These digital distortions are carried out by randomly choosing $\alpha$ for a particular type of distortion (brightness, contrast, color). We have a pre-trained object detection network which could be trained either on the same dataset or any other dataset. 
    \iflogvar
    \begin{figure*}[]
        \centering
        \noindent
            \includegraphics[width=0.095\textwidth]{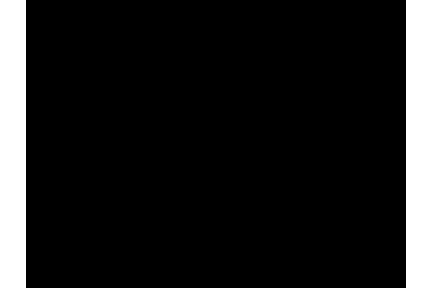}\hspace{0.5mm}%
            \includegraphics[width=0.095\textwidth]{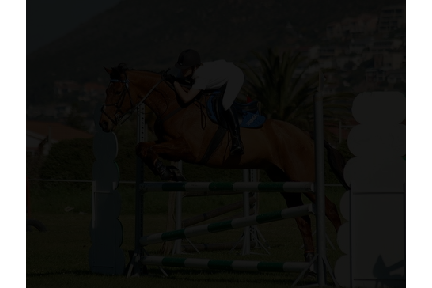}\hspace{0.5mm}%
            \includegraphics[width=0.095\textwidth]{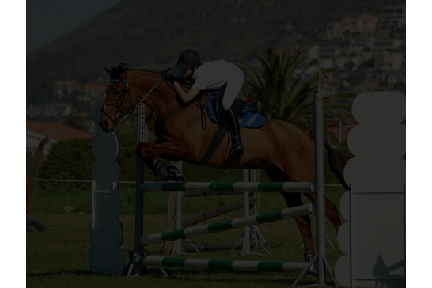}\hspace{0.5mm}%
            \includegraphics[width=0.095\textwidth]{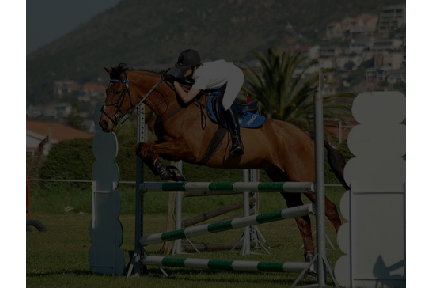}\hspace{0.5mm}%
            \includegraphics[width=0.095\textwidth]{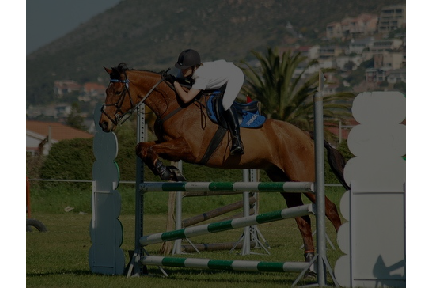}\hspace{0.5mm}%
            \includegraphics[width=0.095\textwidth]{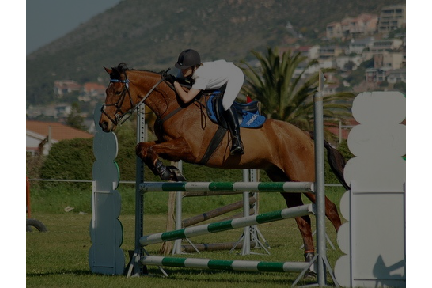}\hspace{0.5mm}%
            \includegraphics[width=0.095\textwidth]{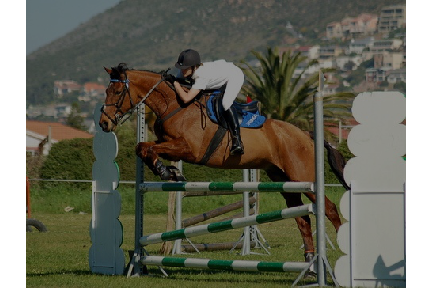}\hspace{0.5mm}%
            \includegraphics[width=0.095\textwidth]{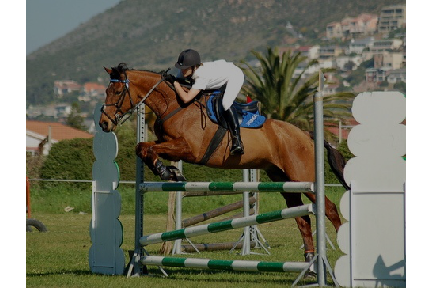}\hspace{0.5mm}%
            \includegraphics[width=0.095\textwidth]{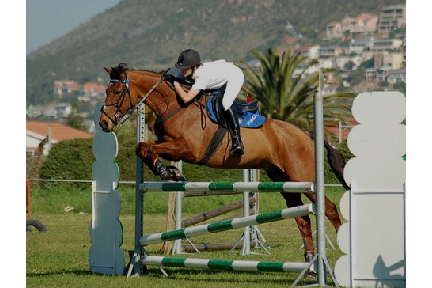}\hspace{0.5mm}%
            \includegraphics[width=0.095\textwidth]{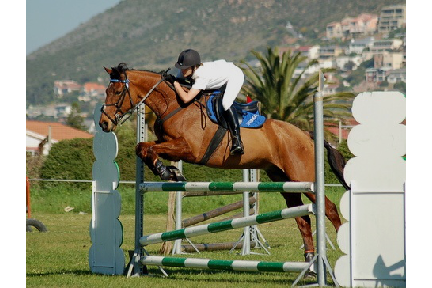}
            \vspace{1mm}
            
            \includegraphics[width=0.095\textwidth]{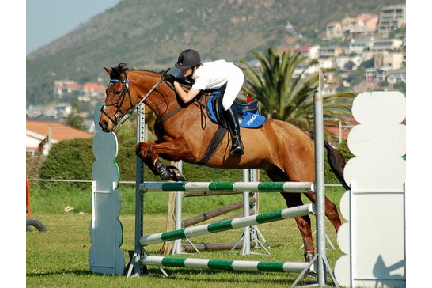}\hspace{0.5mm}%
            \includegraphics[width=0.095\textwidth]{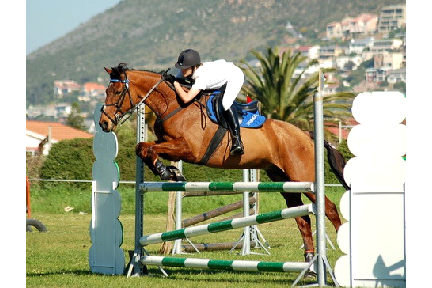}\hspace{0.5mm}%
            \includegraphics[width=0.095\textwidth]{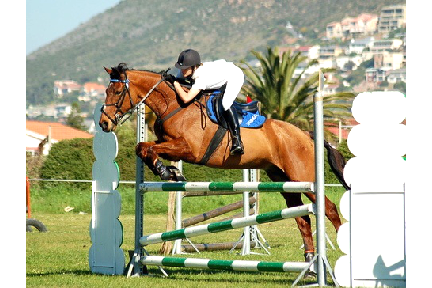}\hspace{0.5mm}%
            \includegraphics[width=0.095\textwidth]{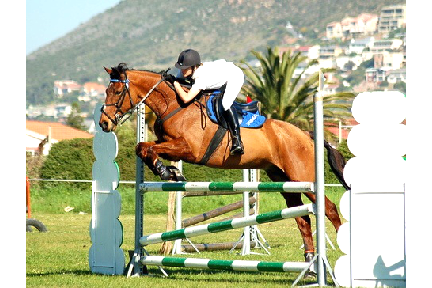} 
            \includegraphics[width=0.095\textwidth]{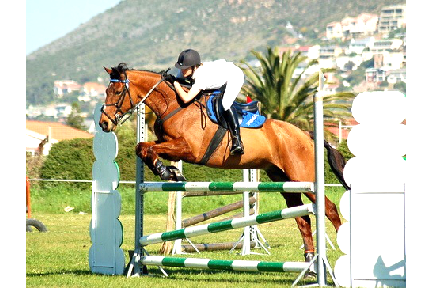}\hspace{0.5mm}%
            \includegraphics[width=0.095\textwidth]{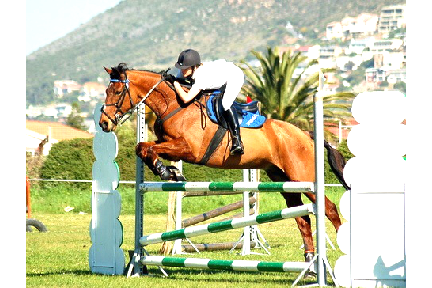}\hspace{0.5mm}%
            \includegraphics[width=0.095\textwidth]{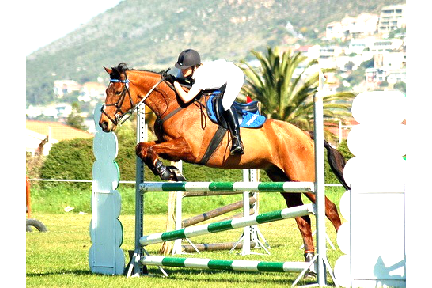}\hspace{0.5mm}%
            \includegraphics[width=0.095\textwidth]{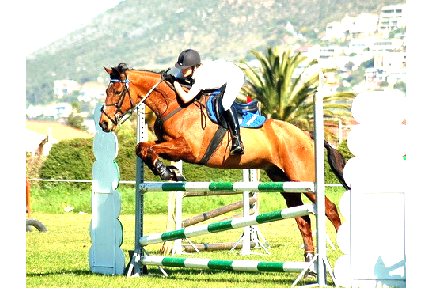}\hspace{0.5mm}%
            \includegraphics[width=0.095\textwidth]{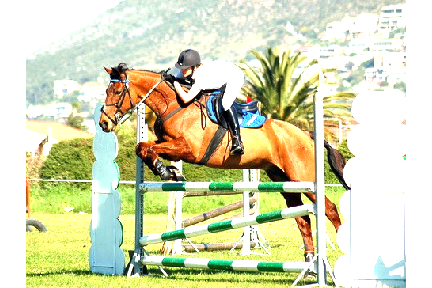}\hspace{0.5mm}%
            \includegraphics[width=0.095\textwidth]{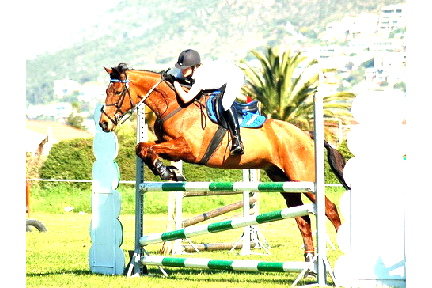}\hspace{0.5mm}%
            \vspace{1mm}
            \par
        \caption{Variation in images with varying brightness distortion factor $\alpha$ from 0 to 2 in steps of 0.1.}
        \label{fig:distortion_range_brightness}
    \end{figure*}
    
    \begin{figure*}[]
        \centering
        \noindent
            \includegraphics[width=0.095\textwidth]{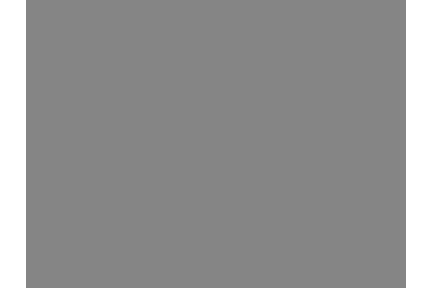}\hspace{0.5mm}%
            \includegraphics[width=0.095\textwidth]{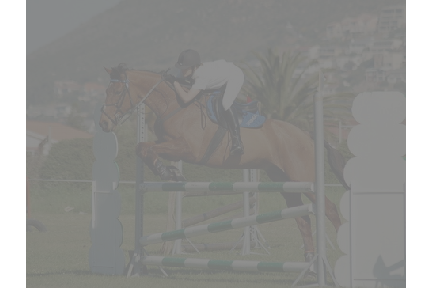}\hspace{0.5mm}%
            \includegraphics[width=0.095\textwidth]{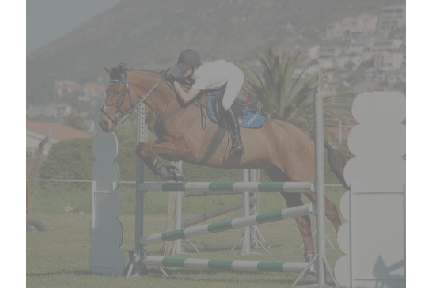}\hspace{0.5mm}%
            \includegraphics[width=0.095\textwidth]{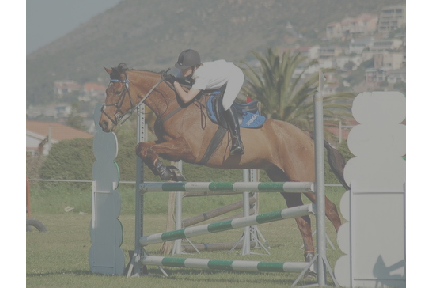}\hspace{0.5mm}%
            \includegraphics[width=0.095\textwidth]{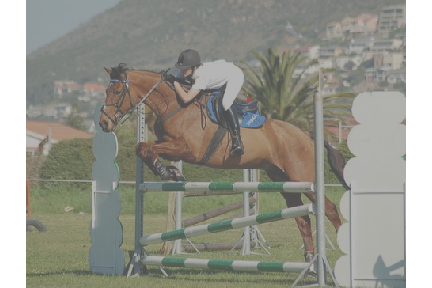}\hspace{0.5mm}%
            \includegraphics[width=0.095\textwidth]{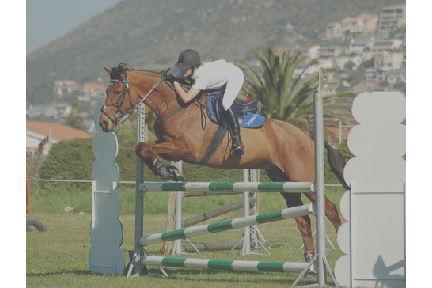}\hspace{0.5mm}%
            \includegraphics[width=0.095\textwidth]{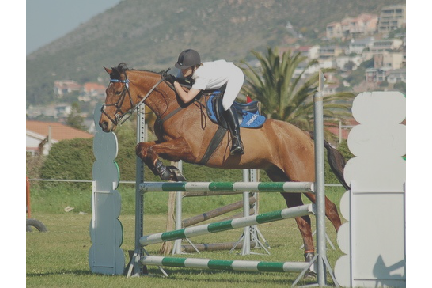}\hspace{0.5mm}%
            \includegraphics[width=0.095\textwidth]{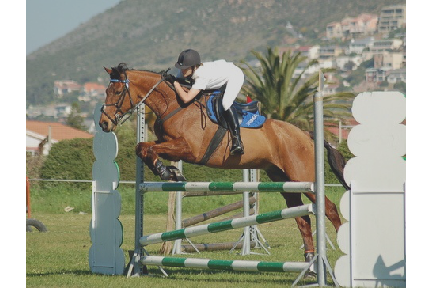}\hspace{0.5mm}%
            \includegraphics[width=0.095\textwidth]{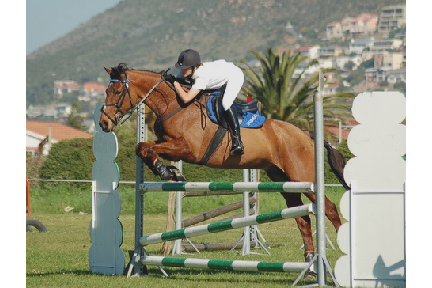}\hspace{0.5mm}%
            \includegraphics[width=0.095\textwidth]{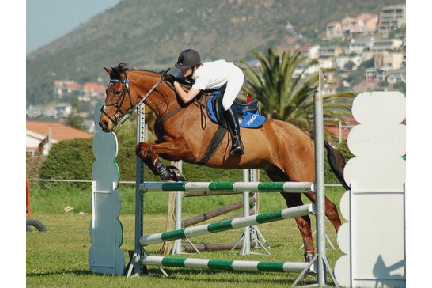}
            \vspace{1mm}
            
            \includegraphics[width=0.095\textwidth]{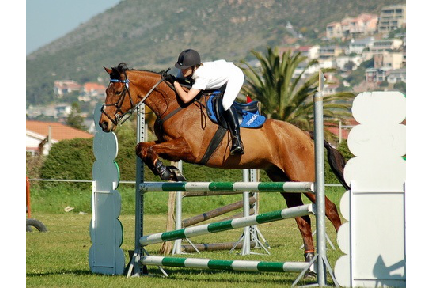}\hspace{0.5mm}%
            \includegraphics[width=0.095\textwidth]{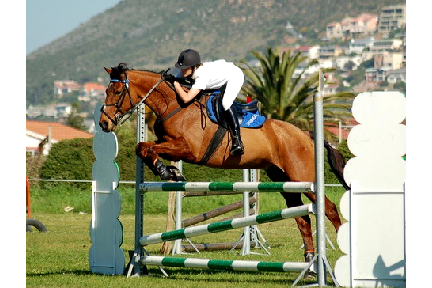}\hspace{0.5mm}%
            \includegraphics[width=0.095\textwidth]{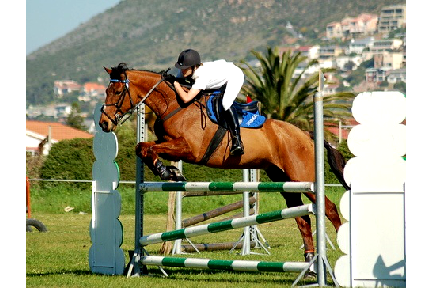}\hspace{0.5mm}%
            \includegraphics[width=0.095\textwidth]{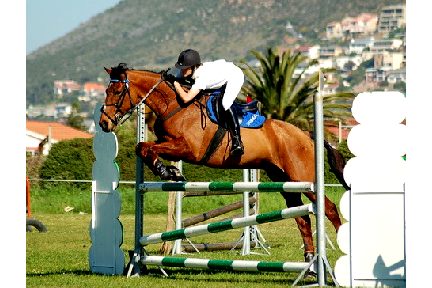} 
            \includegraphics[width=0.095\textwidth]{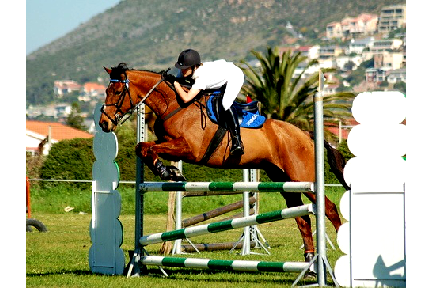}\hspace{0.5mm}%
            \includegraphics[width=0.095\textwidth]{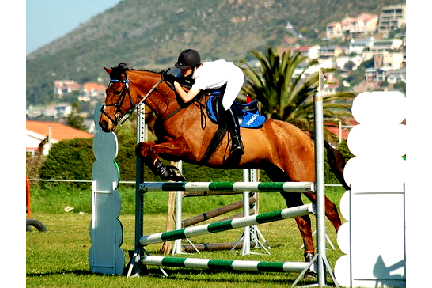}\hspace{0.5mm}%
            \includegraphics[width=0.095\textwidth]{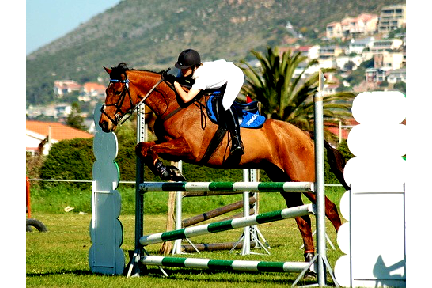}\hspace{0.5mm}%
            \includegraphics[width=0.095\textwidth]{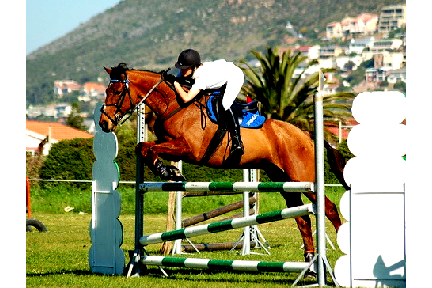}\hspace{0.5mm}%
            \includegraphics[width=0.095\textwidth]{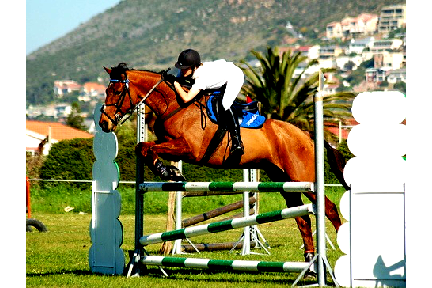}\hspace{0.5mm}%
            \includegraphics[width=0.095\textwidth]{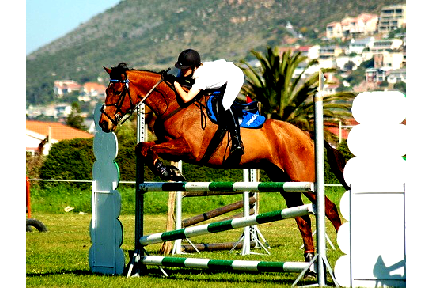}\hspace{0.5mm}%
            \vspace{1mm}
            \par
        \caption{Variation in images with varying contrast distortion factor $\alpha$ from 0 to 2 in steps of 0.1.}
        \label{fig:distortion_range_contrast}
    \end{figure*}
    
    \begin{figure*}[]
        \centering
        \noindent
            \includegraphics[width=0.095\textwidth]{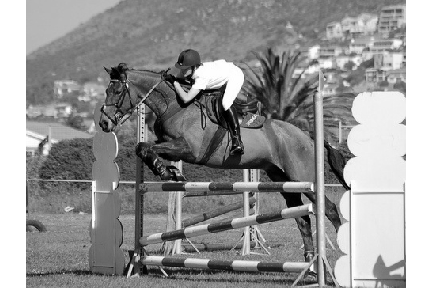}\hspace{0.5mm}%
            \includegraphics[width=0.095\textwidth]{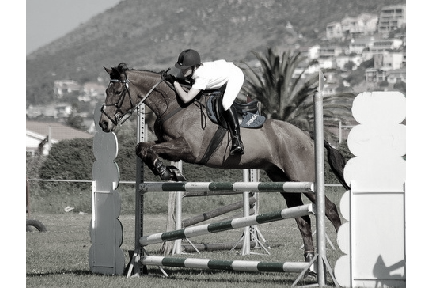}\hspace{0.5mm}%
            \includegraphics[width=0.095\textwidth]{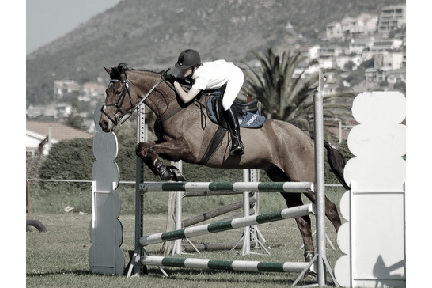}\hspace{0.5mm}%
            \includegraphics[width=0.095\textwidth]{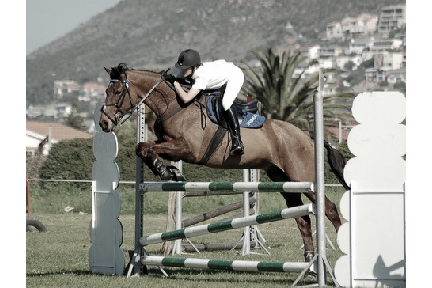}\hspace{0.5mm}%
            \includegraphics[width=0.095\textwidth]{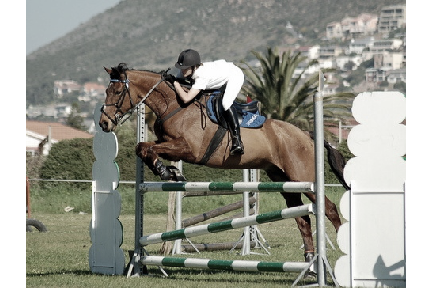}\hspace{0.5mm}%
            \includegraphics[width=0.095\textwidth]{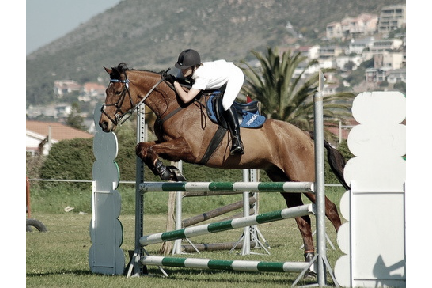}\hspace{0.5mm}%
            \includegraphics[width=0.095\textwidth]{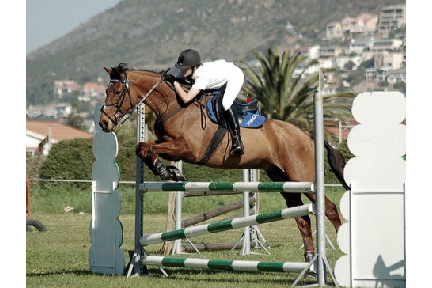}\hspace{0.5mm}%
            \includegraphics[width=0.095\textwidth]{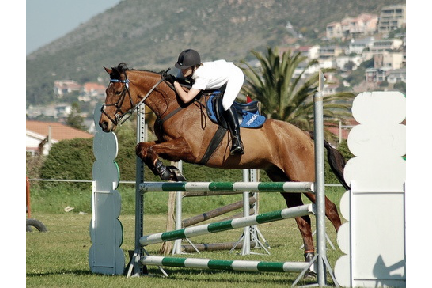}\hspace{0.5mm}%
            \includegraphics[width=0.095\textwidth]{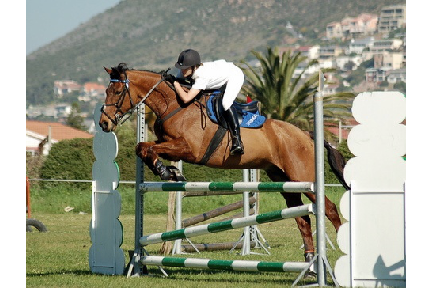}\hspace{0.5mm}%
            \includegraphics[width=0.095\textwidth]{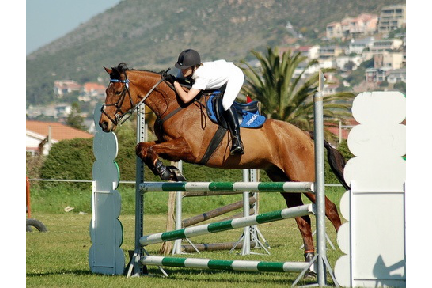}
            \vspace{1mm}
            
            \includegraphics[width=0.095\textwidth]{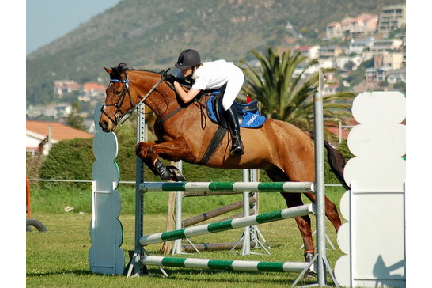}\hspace{0.5mm}%
            \includegraphics[width=0.095\textwidth]{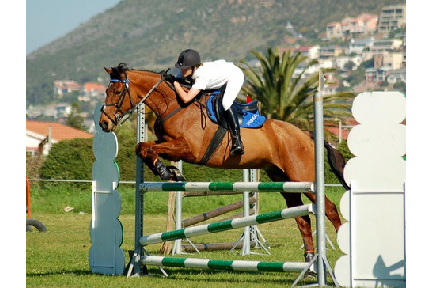}\hspace{0.5mm}%
            \includegraphics[width=0.095\textwidth]{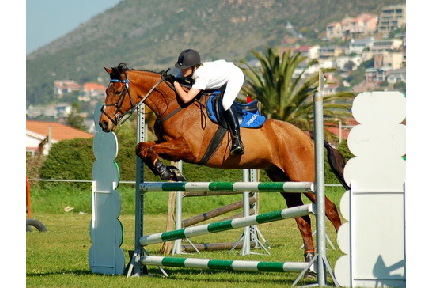}\hspace{0.5mm}%
            \includegraphics[width=0.095\textwidth]{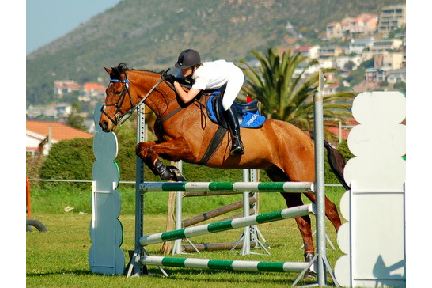} 
            \includegraphics[width=0.095\textwidth]{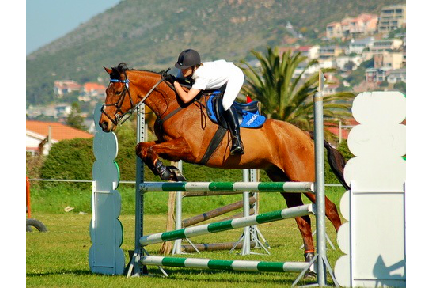}\hspace{0.5mm}%
            \includegraphics[width=0.095\textwidth]{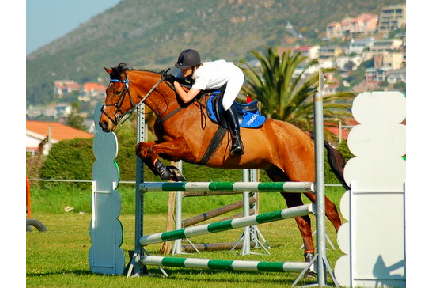}\hspace{0.5mm}%
            \includegraphics[width=0.095\textwidth]{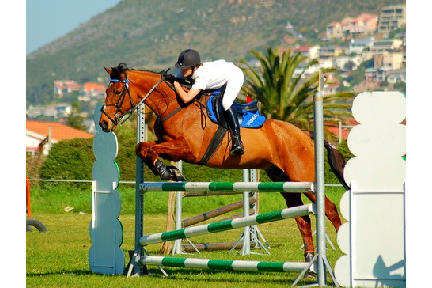}\hspace{0.5mm}%
            \includegraphics[width=0.095\textwidth]{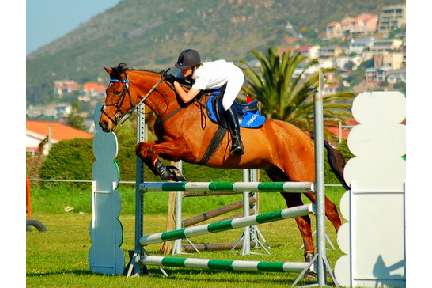}\hspace{0.5mm}%
            \includegraphics[width=0.095\textwidth]{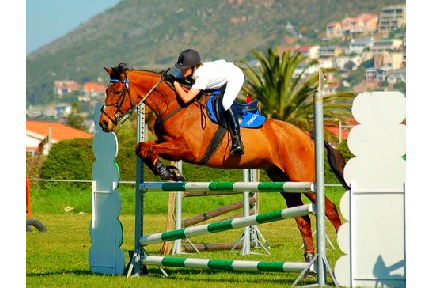}\hspace{0.5mm}%
            \includegraphics[width=0.095\textwidth]{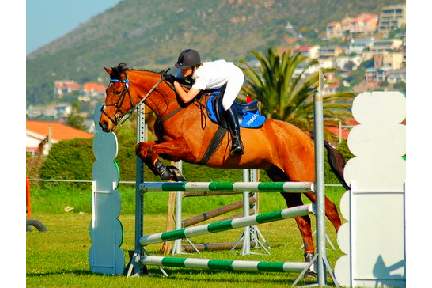}\hspace{0.5mm}%
            \vspace{1mm}
            \par
        \caption{Variation in images with varying color distortion factor $\alpha$ from 0 to 2 in steps of 0.1.}
        \label{fig:distortion_range_color}
    \end{figure*}
    \fi

    \subsection{ObjectRL}\label{section:ObjectRLMDP}
    Formally, the MDP has a set of actions $\mathcal{A}$, a set of states $\mathcal{S}$ and a reward function $\mathcal{R}$ which we define in this section.\\
        \textbf{States}: The states for the agent are  $128\times128\times3$ RGB images from the PascalVOC dataset \cite{Everingham15} which are distorted by random factors $\alpha$ chosen according to the scale of distortion. We consider only one type of distortion (brightness, color, contrast) at a time, ie. we train different models for different types of the distortion. Combining all the different types of distortions in a single model remains to be a key direction to explore in future work.\\
        \textbf{Scales of Distortion:} 
            We perform experiments with the following two degrees of distortion in the image:
            \begin{itemize}
                \item Full-scale distortion: The random distortion in the images $\alpha \in [0,2]$. 
                \item Minor-scale distortion: The random distortion in the images $\alpha \in [0.5,1.8]$. This constraint limits the images to not have distortions which cannot be reverted back with the action space, the agent has access to.
            \end{itemize}
        The variation of the the distorted images can be seen in Fig \ref{fig:distortion_range_brightness}, \ref{fig:distortion_range_contrast}, \ref{fig:distortion_range_color}.\\
        \textbf{Actions}: The agent can choose to change the global parameter (brightness, color, contrast) of the image by giving out a scalar $a_t\in [0,2]$. Here, $a_t$ is equivalent to $\alpha$ in the image distortion equations described in Section \ref{section:img_distortions}. The action $a_t$ can be applied sequentially upto $n$ number of times. After $n$ steps the episode is terminated. Here, we set the value of $n=1$ to achieve stability in training as having larger horizons lead to the images getting distorted beyond repair during the initial stages of learning and hence does not explore with the \textit{better} actions.\\
        \textbf{Reward}: First, we evaluate scores $d_t$ for the images as follows: 
        \begin{equation}
            d_t(x) = \gamma (IoU(x)) + (1-\gamma) (F1(x))
            \label{eqn:reward}
        \end{equation}
        $x$ is the input image to the pre-trained object detector. IoU is the average of all the intersection over union for the bounding boxes predicted in the image and F1 is the F1-score for the image. We set $\gamma=0.1$ because we want to give more importance to the number of correct objects being detected.\\
        
        We evaluate:
        \begin{itemize}
         \setlength\itemsep{-1mm}
            \item $d_{o,t} = d_t(\text{original image})$
            \item $d_{d,t} = d_t(\text{distorted image})$
            \item $d_{s,t} = d_t(\text{state})$
        \end{itemize}
        where the \textit{original image} is the one before the random distortion, \textit{distorted image} is the image after the random distortion and \textit{state} is the image obtained after taking the action proposed by the agent.
        
        We define,
        \begin{equation}
            \beta_t = 2 d_{s,t}-d_{o,t}-d_{d,t}
        \end{equation}
        
        Here, $\beta_t$ is positive if and only if the agent's action leads to an image which gives better detection performance than both the original image as well as the distorted image. Thus we give the reward ($r_t$) as follows:
        \[   
        r_t = 
             \begin{cases}
               \text{+1,} &\quad\text{if } \beta_t \ge -\epsilon \\
               \text{-1,} &\quad\text{otherwise} \\
               
             \end{cases}
        \]
        Note that $d_{o,t} \textrm{ and } d_{d,t}$ do not change in an episode and only $d_{s,t}$ changes over the episode. We set the hyperparameter $\epsilon=0.01$ as we do not want to penalise the minor shifts in bounding boxes which result in small changes in IoU in Eqn[\ref{eqn:reward}]. Fig \ref{fig:flow} shows the training procedure for \textit{ObjectRL}.
        
        \subsection{Motivation for ObjectRL}
        In scenarios where object-detection algorithms are deployed in real-time, for example in autonomous vehicles or drones, lighting conditions and subject speeds can change quickly. If cameras use a single operation mode, the image might be quite blurred or dark and hence the image obtained may not be ideal for performing object detection. In these cases it would not be possible to create new datasets with images obtained from all the possible combinations of camera parameters along with manually annotating them with bounding-boxes. Also, due to the lack of these annotated images we cannot fine-tune the existing object-detection networks on the distorted images. Our model leverages digital distortions on existing datasets with annotations to learn a policy such that it can tackle changes in image parameters in real-time to improve the object detection performance.\\

         One of the main motivations of \textit{ObjectRL} is to extend it to control camera parameters to capture images which are good for object detection in real time. Thus, we propose an extension to \textit{ObjectRL} (for future work) where we have an RL agent which initially captures images by choosing random combinations of camera parameters (exploration phase). A human would then give rewards according to the objects detected in the images in the current buffer. These rewards would then be used to update the policy to improve the choice of camera parameters. This method of assigning a $\{\pm1\}$ reward is comparatively much faster than annotating the objects in the image to extend the dataset and training a supervised model with this extended model. This methodology is quite similar to the DAgger method (Dataset Aggregation) by Ross et al. \cite{DBLP:journals/corr/abs-1011-0686} where a human labels the actions in the newly acquired data before adding it into the experience for imitation learning.


\section{Experiments}\label{section:experiments}
    In this section, we describe the experimental setup for \textit{ObjectRL}. We have built our network with PyTorch. For the object detector, we use a Single Shot Detector (SSD) \cite{DBLP:journals/corr/LiuAESR15} and YOLO-v3 \cite{DBLP:journals/corr/RedmonDGF15} trained on the PascalVOC dataset with a VGG-base network \cite{DBLP:journals/corr/SimonyanZ14a} for SSD.  We use Proximal Policy Optimization (PPO) \cite{DBLP:journals/corr/SchulmanWDRK17} for optimising  the \textit{ObjectRL} agent. We train the agent network on a single NVIDIA GTX 1080Ti with the PascalVOC dataset.
    \\
    
    Both the actor and the critic networks consist of 6 convolutional layers with (kernel size, stride, number of filters)= \{(4,2,8), (3,2,16), (3,2,32), (3,2,64), (3,1,128), (3,1,256)\} followed by linear layers with output size 100, 25, 1. The agent is updated after 2000 steps for 20 epochs with batch-size=64. We use Adam Optimizer \cite{Adam} with a learning rate of $10^{-3}$. We use an $\epsilon-$\textit{Greedy} method for exploration where we anneal $\epsilon$ linearly with the number of episodes until it reaches $0.05$.

\begin{figure*}[t]
    \setlength\tabcolsep{2pt}
    \centering
    \begin{tabular}{ccc}
        \includegraphics[width=0.32\textwidth]{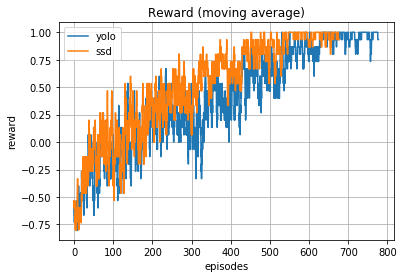}&
        \includegraphics[width=0.32\textwidth]{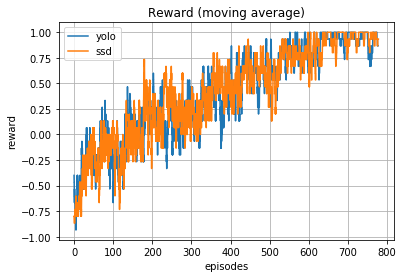}&
        \includegraphics[width=0.32\textwidth]{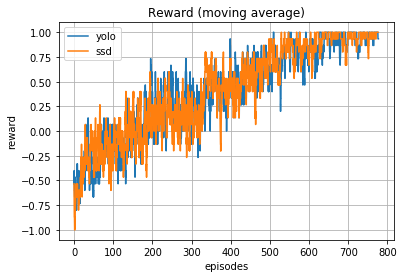}\\
        Brightness & Color & Contrast
    \end{tabular}
    \caption{Episodic return of the \textit{ObjectRL} while training with a moving average of size $30$. Each iteration represents 1K episodes.}
     \label{fig:learning_curves}
\end{figure*}


\section{Results}\label{section:results}
\begin{figure*}[t]
    \setlength\tabcolsep{2pt}
    \centering
    \begin{tabular}{cccc}
        Original&
        \includegraphics[width=0.30\textwidth]{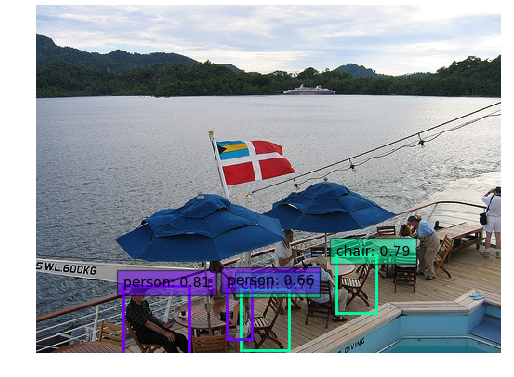}&
        \includegraphics[width=0.30\textwidth]{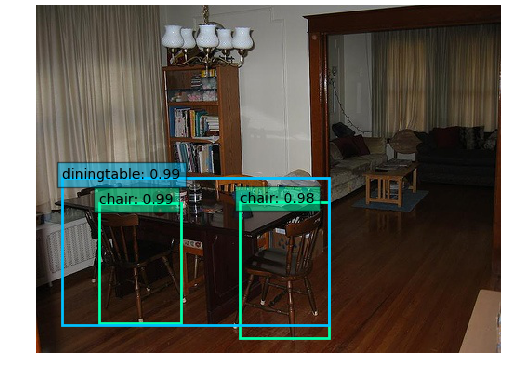}& 
        \includegraphics[height=0.22\textwidth]{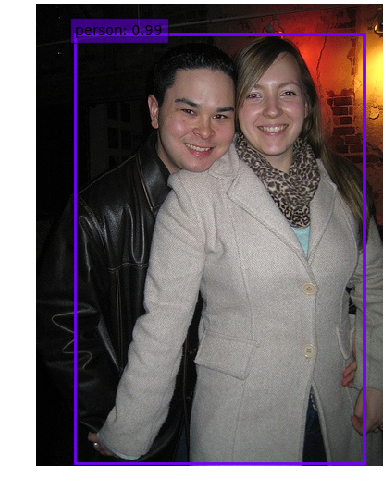}\\
        Distorted&
        \includegraphics[width=0.30\textwidth]{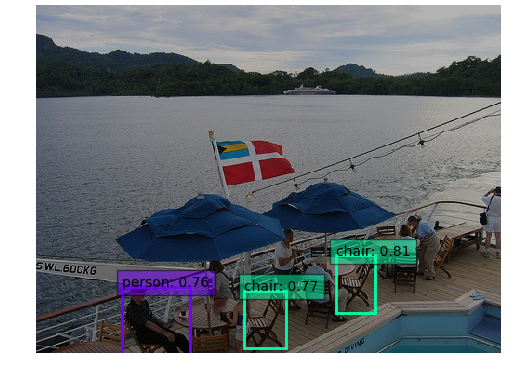}&
        \includegraphics[width=0.30\textwidth]{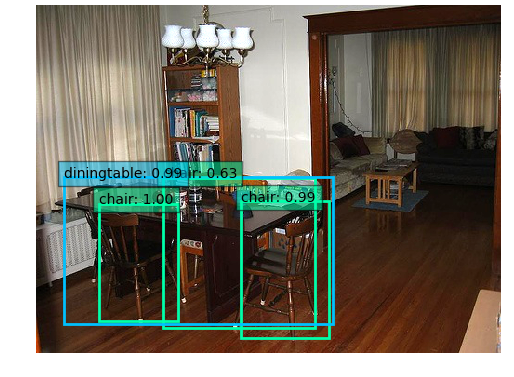}& 
        \includegraphics[height=0.22\textwidth]{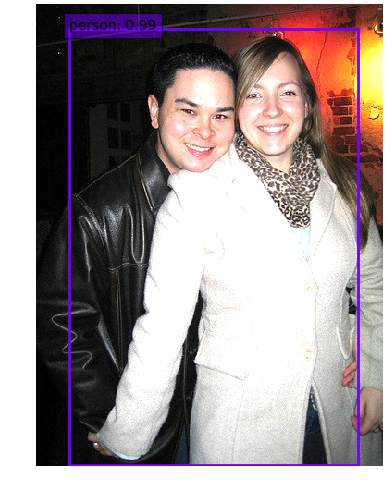}\\
        Agent&
        \includegraphics[width=0.30\textwidth]{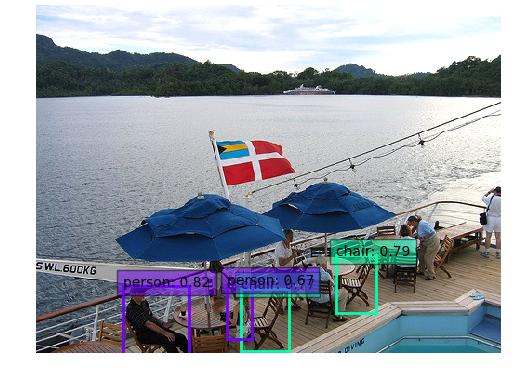}&
        \includegraphics[width=0.30\textwidth]{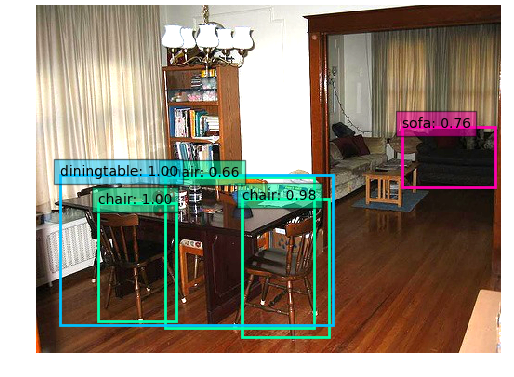}& 
        \includegraphics[height=0.22\textwidth]{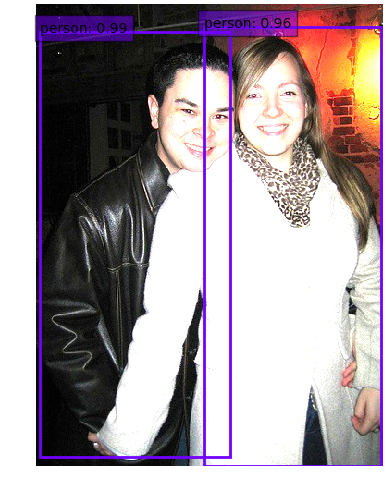}\\
            &
        (a) &
        (b) &
        (c)
    \end{tabular}
    \caption{A few of the outputs from \textit{ObjectRL} with SSD and minor-scale distortion. The top row contains the original images. The second row contains the distorted images. The bottom row contains images obtained from the agent. Bounding boxes are drawn over the objects detected by the detector.}
     \label{fig:objectRL_outputs}
\end{figure*}

\subsection{Measure for evaluation for ObjectRL: TP-Score}
        
        To the best of our knowledge, we believe no suitable measure is defined for this problem and hence we define a measure called \textit{TP-Score(k)} (True Positive Score). This score is the number of images in which $k-$or more true positives were detected which were not detected in the image before transformation. The \textit{TP-Score(k)} is initialised to zero for a set of images $\mathcal{I}$. For example: Let the number of true-positives detected before the transformation be 3 and let the number of true-positives detected after the transformation be 5. Then we have one image where 2 extra true-positives were detected which were not detected in the input image. Thus, we increase \textit{TP-Score(1)} and \textit{TP-Score(2)} by one.

\subsection{Baseline for ObjectRL}
        To obtain the baselines, we first distort the images in the original dataset. The images are distorted with $\alpha$ being randomly chosen from the set $\mathcal{S} = \{0.1,\hdots,1.9,2.0\}$ or $\mathcal{S} = \{0.5,\hdots,1.7, 1.8\}$ depending on the scale.
        The set of available actions to be applied on on these images are: $\hat{\mathcal{S}} = \{\frac{1}{s} \forall s \in \mathcal{S}\}$. We evaluate the \textit{TP-Score(k)} on the distorted images by applying the transformations by performing a grid-search over all $\alpha \in \hat{\mathcal{S}}$ and report the scores obtained with the best-performing actions for different types and scales of distortions in Table \ref{table:objectRL_brightness}, \ref{table:objectRL_color} and \ref{table:objectRL_contrast}. We also report the \textit{TP-Scores} obtained after applying the transformations proposed by \textit{ObjectRL} on the images distorted using full-scale and minor-scales. The scores reported are averaged over 10 image sets $\mathcal{I}$, each containing 10,000 images. Note that the means and standard deviations are rounded to the nearest integers.
        
        

        
        \begin{table}[t]
            \centering
                \begin{tabular}{|c|c|c|c|c|c|c|c|c|} \hline
                    \textbf{k} & \multicolumn{8}{c|}{\textbf{Brightness}}\\ \cline{2-9}
                      & \multicolumn{4}{c|}{Full-scale} 
                      & \multicolumn{4}{c|}{Minor-scale}\\ \cline{2-9}
                      & \multicolumn{2}{c|}{SSD} & \multicolumn{2}{c|}{YOLO} 
                      & \multicolumn{2}{c|}{SSD} & \multicolumn{2}{c|}{YOLO}
                      \\ \cline{2-9}
                     & \textit{GS} & \textit{ObjectRL} & \textit{GS} & \textit{ObjectRL}
                     & \textit{GS} & \textit{ObjectRL} & \textit{GS} & \textit{ObjectRL}\\ \hline\hline
                     1 & $955\pm 14$ & $532\pm20$ & $1360\pm 22$ & $976\pm 18$ 
                       & $435\pm 25$ & $428\pm23$ & $1025\pm 23$ & $883\pm 24$\\ 
                     \hline
                     2 & $154\pm 6$  & $87\pm3$ & $202\pm 15$ & $118\pm15$
                       & $87\pm 12$  & $80\pm9$ & $85\pm 15$ & $63\pm15$\\
                     \hline
                     3 & $49\pm 3$  & $32\pm4$ & $52\pm 8$ & $18\pm 6$
                       & $14\pm 5$  & $12\pm3$ & $8\pm 2$ & $5\pm 1$\\ 
                     \hline
                     4 & $18\pm 3$  & $7\pm1$ & $17\pm 2$ & $4\pm 1$
                       & $5\pm 1$  & $3\pm0$ & $2\pm 0$ & $0$\\ 
                     \hline
                     5 & $7\pm 2$  & $2\pm0$ & $4\pm1$ & $2\pm 0$
                       & $0$  & $0$ & $0$ & $0$\\ [1ex] 
                     \hline
                \end{tabular}
            \caption{\textit{TP-Score(k)} with brightness distortion. GS stands for Grid-Search.}
            \label{table:objectRL_brightness}
        \end{table}

        \begin{table}[t]
            \centering
                \begin{tabular}{|c|c|c|c|c|c|c|c|c|} \hline
                    \textbf{k} & \multicolumn{8}{c|}{\textbf{Color}}\\ \cline{2-9}
                      & \multicolumn{4}{c|}{Full-scale} 
                      & \multicolumn{4}{c|}{Minor-scale}\\ \cline{2-9}
                      & \multicolumn{2}{c|}{SSD} & \multicolumn{2}{c|}{YOLO} 
                      & \multicolumn{2}{c|}{SSD} & \multicolumn{2}{c|}{YOLO}
                      \\ \cline{2-9}
                     & \textit{GS} & \textit{ObjectRL} & \textit{GS} & \textit{ObjectRL}
                     & \textit{GS} & \textit{ObjectRL} & \textit{GS} & \textit{ObjectRL}\\ \hline\hline
                     1 & $973\pm 17$ & $672\pm19$ & $1250\pm 23$ & $1103\pm 21$ 
                       & $561\pm 18$ & $532\pm22$ & $974\pm 21$ & $930\pm 22$\\ 
                     \hline
                     2 & $123\pm 7$  & $84\pm4$ & $210\pm 16$ & $135\pm13$
                       & $43\pm 9$  & $37\pm9$ & $83\pm 12$ & $82\pm12$\\
                     \hline
                     3 & $53\pm 4$  & $31\pm3$ & $63\pm 7$ & $23\pm 6$
                       & $1\pm 0$  & $0$ & $15\pm 2$ & $10\pm 1$\\ 
                     \hline
                     4 & $11\pm 2$  & $3\pm1$ & $19\pm 2$ & $5\pm 1$
                       & $0$  & $0$ & $6\pm 1$ & $3\pm0$\\ 
                     \hline
                     5 & $5\pm 1$  & $1\pm0$ & $6\pm1$ & $2\pm 0$
                       & $0$  & $0$ & $0$ & $0$\\ [1ex] 
                     \hline
                \end{tabular}
            \caption{\textit{TP-Score(k)} with color distortion. GS stands for Grid-Search.}
            \label{table:objectRL_color}
        \end{table}
        
        \begin{table}[t]
            \centering
                \begin{tabular}{|c|c|c|c|c|c|c|c|c|} \hline
                    \textbf{k} & \multicolumn{8}{c|}{\textbf{Contrast}}\\ \cline{2-9}
                      & \multicolumn{4}{c|}{Full-scale} 
                      & \multicolumn{4}{c|}{Minor-scale}\\ \cline{2-9}
                      & \multicolumn{2}{c|}{SSD} & \multicolumn{2}{c|}{YOLO} 
                      & \multicolumn{2}{c|}{SSD} & \multicolumn{2}{c|}{YOLO}
                      \\ \cline{2-9}
                     & \textit{GS} & \textit{ObjectRL} & \textit{GS} & \textit{ObjectRL}
                     & \textit{GS} & \textit{ObjectRL} & \textit{GS} & \textit{ObjectRL}\\ \hline\hline
                     1 & $955\pm 15$ & $532\pm20$ & $1360\pm 21$ & $976\pm 19$ 
                       & $680\pm 22$ & $663\pm24$ & $1038\pm 23$ & $975\pm 24$\\ 
                     \hline
                     2 & $163\pm 8$  & $101\pm4$ & $213\pm 16$ & $134\pm15$
                       & $62\pm 10$  & $49\pm9$ & $104\pm 13$ & $85\pm15$\\
                     \hline
                     3 & $55\pm 4$  & $36\pm4$ & $67\pm 7$ & $39\pm 6$
                       & $14\pm 3$  & $6\pm2$ & $19\pm 3$ & $16\pm 2$\\ 
                     \hline
                     4 & $21\pm 2$  & $11\pm1$ & $28\pm 2$ & $13\pm 1$
                       & $1\pm 0$  & $1\pm0$ & $5\pm 0$ & $3\pm0$\\ 
                     \hline
                     5 & $4\pm 1$  & $2\pm0$ & $5\pm1$ & $2\pm 0$
                       & $0$  & $0$ & $0$ & $0$\\ [1ex] 
                     \hline
                \end{tabular}
            \caption{\textit{TP-Score(k)} with contrast distortion. GS stands for Grid-Search.}
            \label{table:objectRL_contrast}
        \end{table}

    As seen in Table \ref{table:objectRL_brightness},\ref{table:objectRL_color} and \ref{table:objectRL_contrast}, \textit{ObjectRL} is not able to perform as well as the grid-search for full-scale distortions. The reason for this is that many of the images obtained after the full-scale distortions are not repairable with the action set provided to the agent.
    
    But with minor-scale distortions, \textit{ObjectRL} is able to perform as well as the grid-search. The total time taken for the grid-search over all brightness values for one image is $12.5094\pm 0.4103$s for YOLO and $15.1090\pm0.3623$ for SSD on a CPU. The advantage of using \textit{ObjectRL} is that the time taken by the agent is 10 times less than grid-search. This latency is quite crucial in applications like surveillance drones and robots where the lighting conditions can vary quickly and the tolerance for errors in object-detection is low.

    
    \subsection{Discussion on the outputs of \textit{ObjectRL}}
         In this section, we discuss the outputs obtained from \textit{ObjectRL} with SSD and minor-scale distortion which are shown in Fig \ref{fig:objectRL_outputs}. In column (a) 4 true positives are detected in the original image, 3 true positives are detected in the distorted image and 4 true positives are detected in the original image. The distorted image is slightly darker the the original one. \textit{ObjectRL} is able to recover the object lost after distortion. In column (b) 3 true positives are detected in the original image, 4 true positives are detected in the distorted image and 5 true positives are detected in the original image. In this case, even the distorted image performs better than original image. But the agent-obtained image performs the best with 5 true-positives. In column (c) 1 true positive is detected in the original image, 1 true positive is detected in the distorted image and 2 true positives are detected in the original image. In this case the agent obtained image outperforms both the distorted and the original image. For a human eye, the agent-obtained image may not look \textit{pleasing} as it is much brighter than the original image. Ideally for a human, the distorted image in column (c) is the most \textit{pleasing}. Column (c) is one of the perfect examples to demonstrate the fact that whatever looks pleasing to a human eye may not necessarily be the optimal one for object-detection. Thus on an average, the agent is able to recover either as many objects as detected in the original image or more. According to our experiments, there were $8\pm1$ images with SSD and $34\pm5$ images with YOLO-v3, where the agent-obtained image had lesser number of true-positives than the original image. Although, this number of true-positives was more than the number of true-positives detected in the distorted image.
         
         \begin{table}[t]
        \centering
            \begin{tabular}{|c|c|c|c|c|c|c|} \hline
                \textbf{k} & \multicolumn{2}{c|}{\textbf{Brightness}}
                  & \multicolumn{2}{c|}{\textbf{Color}}
                  & \multicolumn{2}{c|}{\textbf{Contrast}}\\  \cline{2-7}
                 & $\pi_{yolo}^{ssd}$
                 & $\pi_{ssd}^{yolo}$
                  & $\pi_{yolo}^{ssd}$ 
                 & $\pi_{ssd}^{yolo}$ 
                  & $\pi_{yolo}^{ssd}$ 
                 & $\pi_{ssd}^{yolo}$ 
                 \\ \hline\hline
                1 & $582\pm 13$ & $1045\pm24$ & $800\pm 15$ & $1249\pm 26$ & $813\pm15$ & $1243\pm26$  \\ 
                 \hline
                 2 & $36\pm 6$  & $73\pm11$ & $72\pm 8$ & $138\pm11$ & $65\pm8$ & $145\pm12$ \\ 
                 \hline
                 3 & $2\pm 0$  & $9\pm4$ & $10\pm1$ & $13\pm3$ & $2\pm0$ & $19\pm4$ \\ [1ex] 
                 \hline
            \end{tabular}
            \caption{\textit{TP-Score(k)} by crossing the policies.}
            \label{table:cross_policies}
        \end{table}
         
    \subsection{Crossing Policies}
        In this section we perform experiments by swapping the detectors for the learned policies. Thus, we use $\pi_{yolo}$ with SSD, (denoted as $\pi_{yolo}^{ssd}$) and $\pi_{ssd}$ with YOLO, (denoted as $\pi_{ssd}^{yolo}$). In Table \ref{table:cross_policies}, we report the number of images where $k-$or lesser true positives were detected with the swapped policy than what were detected using the original policy on their corresponding detectors.
        As shown in Table \ref{table:cross_policies}, $\pi_{SSD}$ on YOLO is worse than $\pi_{YOLO}$ on SSD. This is because the range of values for which SSD gives optimal performance is bigger than the range of values for which YOLO gives optimal performance. In essence, YOLO is more sensitive to the image parameters than SSD.

    
\section{Conclusion}
    This paper proposes the usage of reinforcement learning to improve the object detection of a pre-trained object detector network by changing the image parameters (\textit{ObjectRL}). We validate our approach by experimenting with distorted images and making the agent output actions necessary to improve detection. Our experiments showed that pre-processing of images is necessary to extract the maximum performance from a pre-trained detector. Future work includes combining all the different distortions in a single model and using it for controlling camera parameters to obtain images. Along with this, local image manipulations such as changing the image parameters only in certain regions of the image could be tried out.

\section*{Acknowledgements}
The first author would like to thank Hannes Gorniaczyk, Manan Tomar and Rahul Ramesh for their insights on the project.

%
%
\bibliographystyle{splncs04}
\bibliography{egbib}
\end{document}